\documentclass[11pt,a4paper]{article}
\usepackage[hyperref]{emnlp2020}
\usepackage{times}
\usepackage{latexsym}
\usepackage{graphicx}
\usepackage{booktabs}

\usepackage{microtype}

\usepackage{lingmacros}

\aclfinalcopy % Uncomment this line for the final submission
\title{Multilingual AMR-to-Text Generation}

\author{Angela Fan \\
  FAIR/LORIA \\
  Universit\'{e} de Lorraine, Nancy, France\\
  \texttt{angelafan@fb.com} \\\And
  Claire Gardent \\
  CNRS/LORIA \\
  Nancy, France \\
  \texttt{cgardent@loria.fr} \\}

\date{}

\begin{document}
\maketitle
\begin{abstract}
  Generating text from structured data is challenging because it
  requires bridging the gap between (i) structure and natural language
  (NL) and (ii) semantically underspecified input and fully specified
  NL output. Multilingual generation brings in an additional
  challenge: that of generating into languages with varied word order
  and morphological properties. In this work, we focus on Abstract
  Meaning Representations (AMRs) as structured input, where previous
  research has overwhelmingly focused on generating only into English.
  We leverage advances in cross-lingual embeddings, pretraining, and
  multilingual models to create multilingual AMR-to-text models that
  generate in twenty one different languages. For eighteen languages,
  based on automatic metrics, our multilingual models surpass
  baselines that generate into a single language.  We analyse the
  ability of our multilingual models to accurately capture morphology
  and word order using human evaluation, and find that native speakers
  judge our generations to be fluent.

\end{abstract}

\section{Introduction}

Generating text from structured data has a variety of applications in natural language processing. Tasks such as decoding from tables \cite{lebret2016neural,sha2018order}, question answering from knowledge bases \cite{fan2019using}, and generation from RDF Triples \cite{gardent-etal-2017-webnlg}, knowledge graphs \cite{marcheggiani2018deep} and linguistic meaning representations  \cite{konstas_17} face similar challenges: interpreting structured input and writing fluent output. We focus on generating from graph structures in the form of Abstract Meaning Representations (AMR) \cite{banarescu-etal-2013-abstract}. Previous work has largely focused on generating from AMR into English, but we propose a multilingual approach that can decode into twenty one different languages.  

Compared to multilingual translation, decoding from structured input
has distinct challenges. Translation models take natural language
input and must faithfully decode into natural language
output. However, as shown in \citet{zhao2020bridging}, bridging the gap between 
structured input and linear output is a difficult task. In
addition, in structured input such as graphs, the input is
usually semantically under-specified. For example, in AMRs, function
words are missing and tense and number are not given. Thus, generation
from structured input must bridge the gap between (i) structure and
string and (ii) underspecified input and fully specified
output. Multilinguality brings a third challenge --- that of
generating in languages that have varied morphological and word
order properties.

Annotating natural language with AMR is a complex task and training datasets only exist for English\footnote{AMR datasets from the LDC
  can be found at
  \href{https://amr.isi.edu/download.html}{https://amr.isi.edu/download.html}}, so 
previous work on AMR-to-text generation has overwhelmingly focused on
English. 
We create training data for multilingual AMR-to-Text models, by taking the \textsc{europarl} multilingual corpus 
and automatically annotating the English data with
AMRs using the \texttt{jamr} semantic parser. We then use the English AMRs as
the input for all generation tasks. 
To improve quality, we leverage recent advances in natural language processing such as cross-lingual embeddings, pretraining and multilingual learning.
Cross-lingual embeddings have shown striking improvements on a range
of cross-lingual natural language understanding tasks 
\cite{devlin-etal-2019-bert,conneau2019unsupervised,wu-dredze-2019-beto,pires-etal-2019-multilingual}.
Other work has shown that the pre-training and fine-tuning approaches
also help improve generation performance
\cite{dong2019unified,song2019mass,lawrence-etal-2019-attending,Rothe2019LeveragingPC}.
Finally, multilingual models, where a
single model is trained to translate from multiple source languages
into multiple target languages, are achieving increasingly better 
results in machine translation
\cite{johnson2017google,firat2017multi,aharoni-etal-2019-massively,arivazhagan2019massively}.

By combining these techniques, we demonstrate that fluent generation is
possible for multilingual AMR-to-Text models. We use automatic and human evaluation to assess performance on (1)
\textsc{europarl} data with silver English-centric AMR as input and the 21
\textsc{europarl} languages as target and (2) on \textsc{ldc2015E86} data
with gold English-centric AMR as input and English, Spanish, Italian, and German as
target. Our results
demonstrate, for the first time, that it is possible to generate from
AMRs into multiple languages. 
We show that multilingual models have
strong performance compared to single-language baselines and produce fluent output, based on the judgments of
native speakers. We further investigate how factors, such as differences in the size of the training data,
differences in language word order and morphological properties,
and differences in the set of languages used for training many-to-one
models, impact
results. We will make code and models available, to aid research in multilingual AMR-to-Text
Natural Language Generation.

\begin{figure*}[t]
    \centering
    \includegraphics[width=0.8\linewidth]{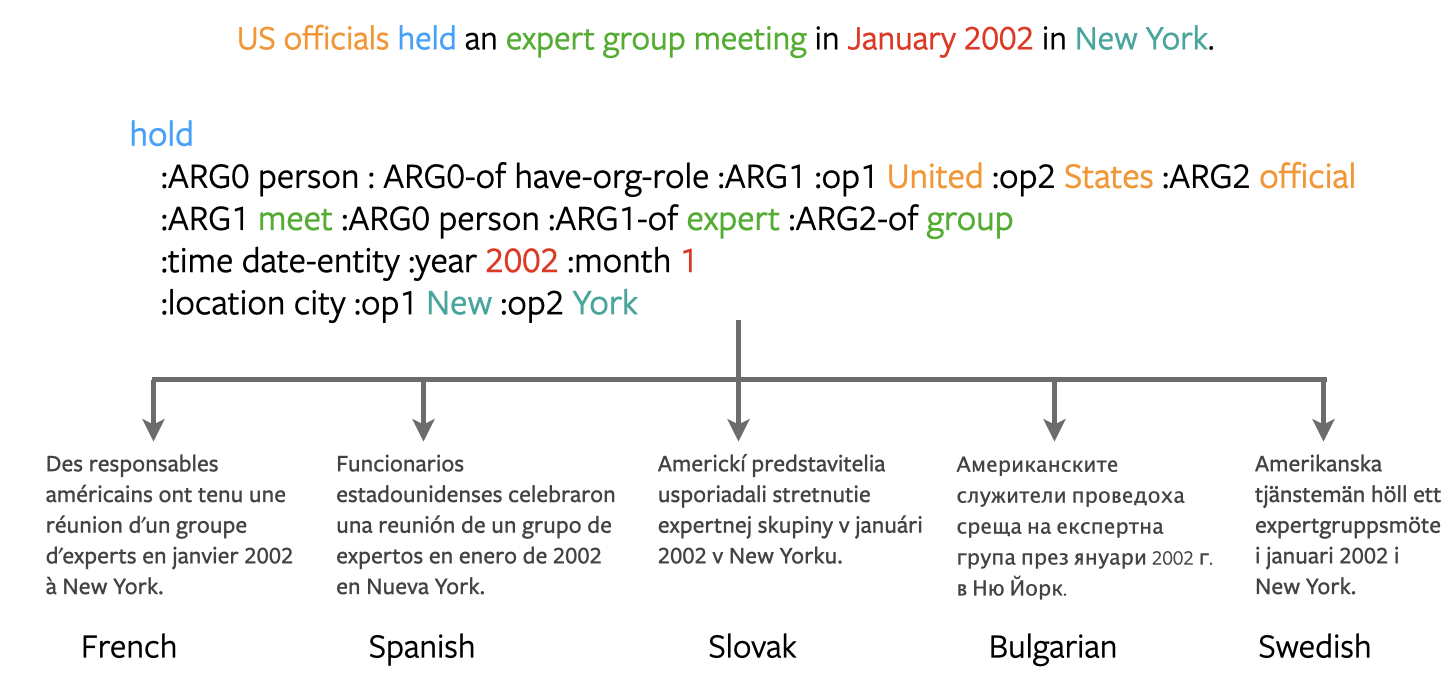}
    \caption{\textbf{Generating into Multiple Languages from English AMR.}}
    \label{fig:pf}
\end{figure*}

\section{Related Work}

\paragraph{AMR-to-Text Generation.}
Initial work on AMR-to-text generation adapted methods from statistical machine translation (MT) \cite{pourdamghani-etal-2016-generating}, grammar-based generation \cite{mille-etal-2017-forge}, tree-to-string transducers \cite{flanigan-etal-2016-generation}, and inverted semantic parsing \cite{lampouras-vlachos-2017-sheffield}. Neural approaches explored sequence-to-sequence models where the AMR is linearized \cite{konstas_17} or modeled with a graph encoder \cite{marcheggiani2018deep,damonte_naacl18,ribeiro-etal-2019-enhancing,song-etal-acl2018,zhu-etal-2019-modeling}. As professionally-annotated AMR datasets are in English, all this work focuses on English. 

One exception is the work of \citet{sobrevilla-cabezudo-etal-2019-back} which uses automatic translation to translate the English text of the LDC AMR data into Brazilian Portuguese and align  English with the Portuguese translation to create Portuguese-centric AMRs. However, this work focuses only on one language. In contrast, we consider generation into twenty one languages. We use very different methods and generate from English-centric AMRs, not target-language AMRs. 

\paragraph{Multilingual MR-to-Text Generation.} While work on AMR-to-Text generation has mostly focused on generation into English, the Multilingual Surface Realization shared tasks \citep{mille2018overview,mille2019overview} have made parallel MR/Text datasets available for 11 languages. 
Two tracks are proposed: a shallow track where the input is an unordered, lemmatized dependency tree and a deep track where the dependency tree edges are labelled with semantic rather than syntactic relations and where function words have been removed.

%However, only a few submissions were truly multilingual
The participants approaches to this multilingual generation task use
gold training data and mostly focus on the shallow track where the
input is an unordered lemmatized dependency tree and the generation
task reduces to linearization and morphological realization. The
models proposed are pipelines that model each of these subtasks and
separate models are trained for each target language
\cite{kovacs-etal-2019-bme,yu-etal-2019-imsurreal,shimorina-gardent-2019-loria,shimorina-gardent-2019-surface,castro-ferreira-krahmer-2019-surface}. In this work, we focus instead on more abstract, deeper, input (AMRs)  and propose
end-to-end, multilingual models for all target languages.

\section{Method}

To generate from AMRs, we use neural sequence to sequence models that model the input AMR with a Transformer Encoder and generate natural language with a Transformer Decoder. For all languages, the input is an English-centric AMR that was derived automatically  using the \texttt{jamr} semantic parser from English text. We pre-train both the AMR encoder and the multilingual decoder and we leverage crosslingual embeddings.

\begin{figure}[t]
    \centering
    \includegraphics[width=0.8\linewidth]{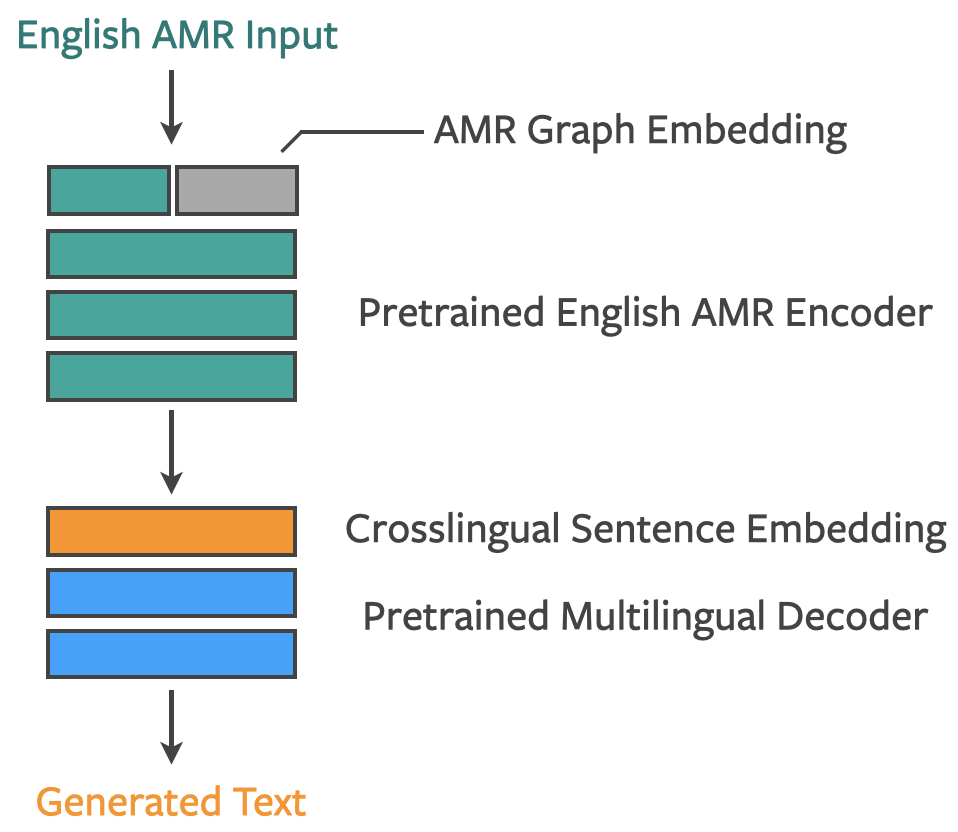}
    \caption{\textbf{One-to-Many Architecture for Multilingual AMR-to-Text Generation.} The English-centric AMR input is linearized and modeled with graph embeddings with a pre-trained Transformer Encoder. Text is generated with a pre-trained Transformer Decoder initialized with cross-lingual embeddings.}
    \label{fig:architecture}
\end{figure}

\subsection{Encoding English AMR}

Abstract Meaning Representations are semantic representations that take the form of a rooted, directed acyclic graph. AMR abstracts away syntax such that sentences with similar meanings have similar AMR graphs. Full detail is not kept by the AMR --- for example, elements such as verb tense are lost. While we focus on decoding from AMR input, the structured form is reflective of other structured inputs used in tasks such as generating from semantic role labels \cite{fan2019strategies} or RDF triples \cite{gardent-etal-2017-webnlg}.

The AMR graph is first linearized into a sequence of tokens as shown in Figure~\ref{fig:pf} after preprocessing following \cite{konstas_17} (see Section 4.1 for a detailed description). Rather than model the graph structure directly, following \citet{fan2019using}, we model the graph using a \textit{graph embedding}. The graph embedding provides additional information to the Transformer Encoder by encoding the depth of each node in the rooted graph and the subgraph each node belongs to. Concretely, each token has a word and position embedding, and additionally an indicator of depth calculated from the root and an indicator of which subtree the node belongs to (with all subtrees stemming from the root).  These additional embeddings are concatenated to the word and position embeddings. Such information allows the Transformer Encoder to capture some graph structure information, while still modeling a sequence. This is depicted in Figure~\ref{fig:architecture}.

To create a one-to-many multilingual model, we model a \textit{language embedding} on the encoder side to allow the decoder to distinguish which language to generate into. This technique has been previously used in multilingual translation \cite{arivazhagan2019massively}. The English AMR begins with a token that indicates the decoder side language. 

To improve the quality of the encoder, we incorporate large-scale pretraining on millions of sequences of AMR by adopting the generative pretraining approach proposed in \citet{lewis2019bart}. This pretraining incorporates various noise operations, such as masking \cite{devlin-etal-2019-bert}, span masking \cite{fan2019using}, and shuffling. Previous work has shown that pretraining is effective for providing neural models with additional information about the structure of natural language and improving model quality \cite{dong2019unified,song2019mass,lawrence-etal-2019-attending}. As models increase in size, smaller training datasets (such as human-annotated AMR) are often not large enough to fully train these models. The entire encoder is pretrained on silver AMRs, as shown in Figure~\ref{fig:architecture}. 

\subsection{Multilingual Decoding from AMR}

The Transformer Decoder attends to the encoded English AMR, a graph of
concepts and relations, and generates text into many different
languages with varied word order and morphology.

As displayed in Figure~\ref{fig:architecture}, we use both language model pretraining and crosslingual embeddings to improve decoder quality.  Monolingual data from various languages is used to pretrain each language model. Further, we incorporate crosslingual embeddings. These embeddings aim to learn universal representations that encode sentences into shared embedding spaces. Various recent work in crosslingual embeddings \cite{conneau2019cross} show strong performance on other multilingual tasks, such as \textsc{xnli} \cite{conneau2018xnli}, \textsc{xtreme} \cite{hu2020xtreme}, and \textsc{mlqa} \cite{lewis2019mlqa}. We use the embeddings from XLM \cite{conneau2019cross} to initialize the multilingual embeddings of our decoder. 

\subsection{Model Training}

To train our one-to-many multilingual AMR-to-text generation model, we use pairs of English AMR and text in multiple different languages. The English AMR does not need to be aligned to sentences in multiple languages. Instead, we create one AMR-to-text corpus for each language and concatenate all of them for training a multilingual model. During the training process, the pretrained AMR encoder and pretrained crosslingual decoder are finetuned on our multilingual AMR-to-text training corpus. 

\section{Experimental Setting}
We describe the various sources of data used to create multilingual AMR-to-text generation models and describe the implementation and evaluation.
\begin{table*}[htbp]
\centering \small
\begin{tabular}{llllllllllll}
\toprule
\bf Model & \bf en & \bf da & \bf de & \bf el & \bf es & \bf fi & \bf fr & \bf it &  \bf nl & \bf pt & \bf sv \\ 
Amount of Data & 8.2M & 1.9M & 1.9M & 1.2M & 1.9M & 1.9M & 2M & 1.9M & 1.9M & 1.9M &  1.9M\\ 
\midrule 
Machine Translation & --- & --- & \bf 17.8 & --- & 24.9 & --- & \bf 20.7 & 18.6 & 19.4 & 21.0 & 19.2 \\
\midrule 
English AMR-XX & \bf 34.2 & 21.3 & 16.9 & 14.2 & 24.3 & 12.9 & 20.5 & \bf 19.1 & 18.8 & 20.4 & 18.6 \\ 
\midrule
Multilingual & 32.5 & 21.2 & 17.0 & 13.8 & 24.2 & 12.4 & 19.7 & 17.8 & 18.5 & 20.5 & 18.7 \\
+ Graph Embedding & 32.9 & 21.4 & 17.0 & 14.0 & 24.3 & 12.5 & 19.9 & 18.0 & 18.6 & 20.7 & 18.9 \\
+ Crosslingual Embedding & 33.0 & 21.7 & 17.3 & 14.4 & 24.7 & 12.9 & 19.9 & 18.5 & 19.0 & 21.0 & 19.0 \\
+ Encoder Pretraining & 33.4 & 21.7 & 17.3 & 14.5 & 24.9 & 13.0 & 20.2 & 18.7 & 19.1 & 21.0 & 19.1 \\
+ Decoder Pretraining & 33.8 & \bf 21.9 & 17.5 & \bf 14.6 & \bf 25.1 & \bf 13.4 & 20.3 & 18.9 & \bf 19.4 & \bf 21.2 & \bf 19.5 \\ 
\bottomrule
% \bottomrule
\toprule
\bf Model & & \bf bg & \bf cs & \bf et & \bf hu & \bf lt & \bf lv & \bf pl & \bf ro & \bf sl & \bf sk \\ 
Amount Data & & 400K & 650K & 650K & 620K & 630K & 640K & 630K & 400K & 640K & 620K \\
% \midrule 
% Machine Translation & \\
\midrule 
English AMR-XX & & 33.8 & 27.5 & 18.9 & 23.1 & 23.9 & 25.4 & 23.4 & 30.6 & 30.1 & 28.7 \\ 
\midrule
Multilingual & & 34.6 & 28.4 & 19.1 & 23.8 & 24.4 & 26.9 & 23.4 & 31.5 & 30.6 & 29.7 \\
+ Graph Embedding & & 34.7 & 28.5 & 19.3 & 23.9 & 24.5 & 27.0 & 23.6 & 31.5 & 20.7 & 29.9 \\
+ Crosslingual Embedding & & 35.0 & 28.9 & 19.7 & 24.3 & 24.8 & 27.4 & 24.0 & 31.7 & 30.8 & 30.1 \\
+ Encoder Pretraining & & 35.2 & 29.0 & 19.8 & 24.5 & 25.0 & 27.5 & 24.1 & 31.9 & 31.0 & 30.2 \\
+ Decoder Pretraining & & \bf 35.7 & \bf 29.5 & \bf 21.2 & \bf 24.7 & \bf 25.5 & \bf 27.9 & \bf 24.4 & \bf 32.1 & \bf 31.4 & \bf 30.6 \\
\bottomrule
\end{tabular}
\caption{\label{table:full_results} \textbf{Results on 21 Languages in Europarl}. The English-XX baseline (generation into a single language) combines all modeling improvements. When training on multiple seeds, the standard deviation is around 0.1 to 0.3 BLEU, making the difference between the multilingual baseline and the addition of our modeling improvements statistically significant.}
\end{table*}

%\vspace{-1cm}

\subsection{Data}

\paragraph{Pretraining} For encoder pretraining on silver AMR, we take thirty million sentences from the English portion of \textsc{ccnet}\footnote{\url{https://github.com/facebookresearch/cc_net}} \cite{wenzek2019ccnet}, a cleaned version of Common Crawl (an open source version of the web). We use \texttt{jamr}\footnote{\url{https://github.com/jflanigan/jamr}} to parse English sentences into  AMR. 
For multilingual decoder pretraining, we take thirty million sentences from each language split of \textsc{ccnet}. 

\paragraph{Multilingual Data} We use \textsc{europarl}, an aligned corpus of European Union parliamentary debates. Each language in \textsc{europarl} is aligned to English. We study the twenty one languages available in \textsc{europarl}: Bulgarian, Czech, Danish, Dutch, English, German, Greek, Spanish, Estonian, Finnish, French, Hungarian, Italian, Lithuanian, Latvian, Polish, Portuguese, Romanian, Slovak, Slovenian, and Swedish. The earliest releases in \textsc{europarl} were prepared with a fixed common testing set across all languages, but later releases in ten new languages do not have a validation or test set. Thus, for the languages where the standard split is applicable, we report results on the common testing set, splitting it in half for validation and testing. For languages where there is no evaluation set, we take a part of the training set and reserve it for validation and another portion for testing. We use \texttt{jamr} to parse the English text of the Europarl corpus into AMRs. This creates a corpus of automatically created silver English AMRs aligned with sentences in twenty one European languages. 

\paragraph{Gold AMR} 
We also evaluate our models (trained on silver AMRs) on gold AMR where available. For this, we use the \textsc{crosslingual amr} dataset from \citet{cohen2018cross}\footnote{To evaluate on this data, please contact \citet{cohen2018cross}}.
The corpus was constructed by having professional translators translate the English text of the \textsc{ldc2015E86} test set into Spanish, Italian, German, and Chinese. We only evaluate on languages where we have training data from \textsc{europarl} (i.e. we do not include Chinese as it is not in \textsc{europarl}).

\paragraph{Preprocessing}
All data remains untokenized and cased. For AMR, we follow \citet{konstas_17} in processing the \texttt{jamr} output into a simpler form. We remove variable names and instance-of relation ( / ) before every concept. However, we do not anonymize entities or dates, as improvements in modeling have allowed for better representations of rare words such as entities. We learn a sentencepiece model with 32K operations to split the English AMR into subword units. On the decoder side, we apply the sentencepiece model and vocabulary of XLM \cite{conneau2019cross}. We choose to use the existing XLM sentencepiece and vocabulary so that the XLM cross-lingual embeddings can be used to initialize our models. For the encoder, we do not use existing vocabularies, as they do not capture the AMR graph structure. 
%\vspace{-0.2cm}
\subsection{Models}
We implement our models in \texttt{fairseq-py} \cite{ott2019fairseq}. We use large Transformer \cite{vaswani2017attention} sequence-to-sequence models and train all models for 50 epochs with LayerDrop \cite{fan2019reducing}, which takes around 2 days. We initialize all weights with the pretrained models. When combining crosslingual word embeddings and encoder and decoder pretraining, we initialize all weights with pretraining, then use  crosslingual word embeddings. We do not perform extensive hyperparameter search, but experimented with various learning rate values to maintain stable training with pretrained initialization. To generate, we decode with beam search with beam size 5. Our pretrained models are available for download.\footnote{\url{https://github.com/facebookresearch/m-amr2text}}
\subsection{Monolingual and Translation Baselines}
We compare our multilingual models both to monolingual models (one
model trained for each language) and to a hybrid NLG/MT baseline. For
the latter, we first generate with the AMR-to-English model and then
translate the generation output to the target language using MT. Our
translation models are Transformer Big models trained with LayerDrop
\cite{fan2019reducing} for 100k updates on public benchmark data from
WMT where available and supplemented with mined data from the
\texttt{ccMatrix} project \cite{schwenk2019ccmatrix}. We trained translation models for languages where large quantities of aligned bitext data are readily available, and cover a variety of languages.

\begin{table}[!h]
\centering\small
\begin{tabular}{lllll}
\toprule
\bf Model & \bf en & \bf es & \bf it & \bf de \\  
\midrule
Konstas et al. (2017) & 22.0 & --- & --- & --- \\ 
Song et al. (2018) & 23.3 & --- & --- & --- \\ 
Cao et al. (2019) & 23.5 & --- & --- & --- \\ 
Damonte et al. (2019) & 24.4 & --- & --- & --- \\ 
Guo et al. (2019) & 25.7 & --- & --- & --- \\ 
Ribeiro et al. (2019) & 24.3 & --- & --- & --- \\ 
Zhu et al. (2019) & \bf 29.7 & --- & --- & --- \\ 
\midrule
Machine Translation & --- & 21.6 &  19.6 & \bf 15.7 \\ 
\midrule 
English-XX Seq2Seq & 25.2 & 21.1 & \bf 19.8 & 14.9 \\ 
\midrule
Multilingual Seq2Seq & 24.2 & 21.0 & 19.0 & 14.7 \\ 
+ Graph Attribute & 24.5 & 21.0 & 19.2 & 14.8 \\ 
+ Crosslingual Embed & 24.6 & 21.3 & 19.4 & 15.1 \\ 
+ AMR Enc Pretrain & 24.7 & 21.5 & 19.6 & 15.1 \\ 
+ Multiling Dec Pretrain & 24.9 & \bf 21.7 & \bf 19.8 & 15.3 \\ 
+ Finetune on Gold AMR & 26.3 & --- & --- & --- \\ 
\bottomrule
\end{tabular}
\caption{\label{table:gold_results} \textbf{Results on Gold AMR from LDC2015E86}.}
\end{table}

\begin{table*}[htbp]
\centering\small
\begin{tabular}{lccccc|cc}
\toprule
 & \bf Number of & & \bf Word & \bf Semantic & \bf Good & \bf Std Dev & \bf Std Dev \\ 
\bf Language & \bf Evaluators & \bf Morphology & \bf Order &  \bf Accuracy & \bf Paraphrases & \bf Morphology & \bf Word Order \\  
\midrule
English & 7 & 2.9 & 2.9 & 2.4 & 84\% & 0.06 & 0.04 \\
Danish & 2 & 2.9 & 2.9 & 2.3 & 88\% & 0.09 & 0.04 \\ 
German & 4 & 3.0 & 2.9 & 2.2 & 75\% & 0.02 & 0.06 \\ 
Greek & 5 & 2.9 & 2.9 & 2.2 & 75\% & 0.06 & 0.04 \\ 
Spanish & 10 & 2.9 & 2.9 & 2.2 & 81\% & 0.09 & 0.07 \\ 
Finnish & 2 & 2.9 & 3.0 & 2.1 & 69\% & 0.01 & 0.00 \\ 
French & 7 & 3.0 & 3.0 & 2.3 & 81\% & 0.02  & 0.03  \\ 
Italian & 5 & 3.0 & 3.0 & 2.3 & 82\% & 0.04 & 0.05 \\ 
Dutch & 7 & 2.9 & 2.9 & 1.9 & 60\% & 0.06 & 0.06 \\ 
Portuguese & 7 & 2.9 & 2.9 & 2.4 & 83\% & 0.08 & 0.06 \\
Swedish & 5 & 2.9 & 2.9 & 2.3 & 84\% & 0.04 & 0.08 \\ 

Bulgarian & 6 & 2.8 & 2.8 & 2.0 & 67\% & 0.07 & 0.11 \\ 
Czech & 3 & 2.9 & 2.8 & 2.3 & 79\% & 0.05 & 0.11 \\ 
Estonian & 1 & 2.9 & 2.9 & 2.2 & 78\% & --- & --- \\ 
Hungarian & 5 & 2.6 & 2.5 & 2.1 & 70\% & 0.14 & 0.23 \\ 
Latvian & 3 & 2.8 & 2.7 & 2.1 & 74\% & 0.07 & 0.16 \\ 
Polish & 2 & 2.8 & 2.9 & 1.6 & 54\% & 0.10  & 0.04\\ 
Romanian & 10 & 2.7 & 2.7 & 1.9 & 68\% & 0.22 & 0.23 \\ 
\bottomrule
\end{tabular}
\caption{\label{table:human_eval} \textbf{Human Evaluation}. Native speakers assess fifty sentences on a scale of 1 to 3, with 3 the highest score. Good Paraphrases are sentences with high scores (2 or 3) for both Semantic Accuracy and Paraphrasing.}
\end{table*}

\subsection{Evaluation} We evaluate with detokenized BLEU using \texttt{sacrebleu} \cite{post-2018-call}. We conduct human evaluation by asking native speakers to evaluate word order, morphology, semantic faithfulness (with respect to the reference) and paraphrasing (how much the generation differs from the reference) on a 3 point scale. The evaluation was done online. For each language, evaluators annotated 25 test set sentences with high BLEU score and 25 sentences with low BLEU score. We removed sentences that were shorter than 5 words. As it is difficult to ensure high quality annotations for 21 languages using crowdsourcing, we relied on colleagues by reaching out on NLP and Linguistics mailing lists. As a result, the number of evaluators per language varies (cf. Table \ref{table:human_eval}).

We evaluate multilingual AMR-to-Text generation models in 21
languages. We conduct an ablation study which demonstrates the
improvements in modeling performance induced by incorporating graph
embeddings, cross lingual embeddings, and pretraining. Finally, we
analyze model performance with respect to several linguistic attributes (word order, morphology, paraphrasing, semantic faithfulness) using both automatic metrics and human evaluation.

\subsection{Multilingual AMR-to-Text Generation}

\paragraph{Monolingual vs. Multilingual Models.}
We compare English-XX baselines trained to generate from AMR into a single language with multilingual models. We note that as the English-XX models specializes for each language, they have less to model with the same parameter capacity.
Results are shown in Table~\ref{table:full_results}. 
Overall, multilingual models perform well --- on 18 of the 21 languages, the performance measured by BLEU is stronger than the monolingual baseline.

One advantage of multilingual AMR-to-Text generation is increased quantities of AMR on the encoder side. This is particularly helpful when the size of the training data is low. For instance, Estonian (\textit{et}) sees a 2.3 BLEU point improvement from multilingual modeling.
Conversely, languages such as English, Swedish and French
benefit less from multilingual modeling, most likely because there is sufficient data for those languages already. 
More generally, there is a marked difference between languages for which the training data is large and those for which the training data is smaller. When the training data is large (1.9 to 2M training instances, top part of Table~\ref{table:full_results}), the average improvment is +0.36 BLEU (Min:-0.2, Max:+0.9) whereas for languages with smaller training data (400 to 620K training instances, bottom part of Table~\ref{table:full_results}\footnote{For many languages, such as Slavic languages, it is because the EU expanded to include these countries later on. Thus there is less European Parliamentary proceeding data.}), the average improvement is +1.75 (Min:+1, Max:+2.3).  These trends are similar to observations on other tasks --- namely that pretraining is most helpful when there is not sufficient training data in the task itself to train strong representations. 

\section{Results}

\paragraph{Performance on Gold English AMR} 
We evaluate our models trained on silver AMR on the 
\textsc{crosslingual amr} dataset from \citet{cohen2018cross} where the input is a gold English-centric AMR and the output is available in three European languages: Spanish, French, and Italian.
The results are shown in Table~\ref{table:gold_results}. Similar to the trends seen when generating from silver AMR, we find that multilingual models have strong performance. BLEU scores are lower than on \textsc{europarl} as the models are tested out of domain (training on parliamentary debates but testing on newswire and forum data domains). 

On English LDC data, we compare to existing work. Even though it is trained on silver AMRs and out of domain, non-LDC data, the multilingual model compares well with previous work (see Table~\ref{table:gold_results}).
When finetuned on the \textsc{ldc2015E86} train set, our model improves on English by over 1 BLEU point, outperforming all previous work except \citet{zhu-etal-2019-modeling}. This work directly models the graph structure of AMR with structure aware attention to improve Transformer architectures --- this is orthogonal to our main aim of multilingual generation and can be incorporated in future work.

\paragraph{Impact of Modeling Improvements.} For the multilingual model, we display the effect of incrementally adding additional modeling improvements (cf. Table~\ref{table:full_results}). Each improvement is essentially universally helpful across all considered languages, though some have a greater improvement on performance than others. 

\begin{table*}[htbp]
\centering\small
\begin{tabular}{lccc}
\toprule
\bf Evaluation & \bf Morphology & \bf Word Order & \bf Semantic Accuracy \\ 
\midrule
\bf Spanish & \\ 
Machine Translation & 2.9 & 2.7 & 2.0 \\ 
Multilingual AMR to Text & 2.8 & 2.9 & 2.1 \\
\midrule 
\bf Italian & \\ 
Machine Translation & 3.0 & 2.9 & 2.2 \\ 
Multilingual AMR to Text & 2.9 & 3.0 & 2.1 \\
\midrule 
\bf German & \\ 
Machine Translation & 2.8 & 2.9 & 2.0 \\ 
Multilingual AMR to Text & 3.0 & 3.0 & 2.2 \\
\bottomrule
\end{tabular}
\caption{\label{table:human_eval_mt} \textbf{Human Evaluation of our approach compared to the Hybrid English AMR to Text + Machine Translation baseline using Gold AMR from LDC2015E86}. Two native speakers per language assess fifty sentences each on a scale of 1 to 3, with 3 being the highest score.}
\end{table*}
\begin{table*}[htbp]
\centering \small
\begin{tabular}{lll}
\toprule
\bf English & \bf Generation & This point will certainly be the subject of subsequent further debates in the council. \\
& \bf Reference & This is a point that will undoubtedly be discussed later in the Council.
\\
\\
\bf French & \bf Generation & Je ne suis pas favorable \`{a} des exceptions  \`{a} cette r\`{e}gle.
\\
& \bf Reference & A mon avis, il n'est pas bon de faire des exceptions  \`{a} cette r\`{e}gle.
\\
\\
\bf Swedish & \bf Generation & Därför röstade vi inte för detta betänkande.
\\
& \bf Reference & Vi har därför inte röstat för detta betänkande. 
\\
\bottomrule
\end{tabular}
\caption{\label{table:generation_examples} \textbf{Example Paraphrases} generated by our multilingual model.}
\end{table*}

\paragraph{Comparison to the Hybrid NLG/MT Baseline.}
Compared to the NLG/MT baseline, our multilingual models provide comparable results while providing an arguably simpler approach (end-to-end rather than pipeline) and training on much lower quantities of parallel data --- on German and French (very high resource languages with millions of examples of training data), there is slightly stronger performance. On other languages we compare to, the translation models perform a bit worse.
\begin{table}[htbp]
\centering\small
\begin{tabular}{llllll}
\toprule
\bf Model & \bf es & \bf fr & \bf it & \bf pt & \bf ro \\  
One Language & 25.2 & 20.3 & 18.9 & 22.2 & 32.1 \\ 
Romance Family & 25.5 & 20.5 & 19.3 & 22.5 & 32.5 \\ 
All Languages & 25.3 & 20.5 & 19.3 & 22.4 & 32.2 \\ 
\midrule 
 &  &  \bf da & \bf de & \bf nl & \bf sv \\  
One Language & & 21.3 & 17.0 & 18.5 & 18.7 \\ 
Germanic Family & & 21.8 & 21.9 & 19.6 & 19.3 \\ 
All Languages & & 21.9 & 17.5 & 19.4 & 19.5 \\ 
\bottomrule
\end{tabular}
\caption{\label{table:language_families} \textbf{Performance when training with increasingly more languages.} Training one multilingual AMR-to-Text model with languages in the related language family improves performance.}
\end{table}

% this is annoying because we discuss it BEFORE our other human eval experiments
We further conduct a human evaluation study on Spanish, Italian, and German. We ask evaluators to assess the morphology, word order, and semantic accuracy of our Multilingual AMR to Text system compared to this hybrid English AMR to Text + Machine Translation baseline. We show in Table~\ref{table:human_eval_mt} that the two models score very similarly in human evaluation, indicating the strength of this fully multilingual system in producing fluent output.

\subsection{Analysis of Multilingual Generation}

A core challenge for multilinguality is that languages
differ with respect to word order and morphology, so models must learn this per language. We use 
automatic and human evaluation to investigate how these
differences affect performance.

\paragraph{Morphology}

\begin{figure}[ht]
    \centering
    \includegraphics[width=\columnwidth]{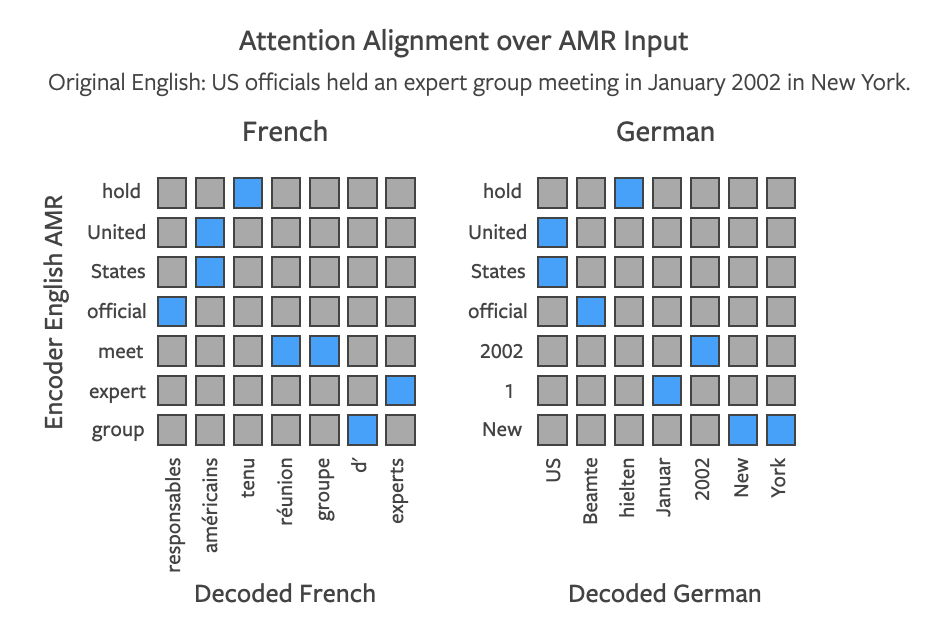}
    \caption{\textbf{Attention alignment} when decoding in French and German from the same input AMR.}
    \label{fig:alignment}
\end{figure}
\begin{figure}[ht]
    \centering
    \includegraphics[width=\columnwidth]{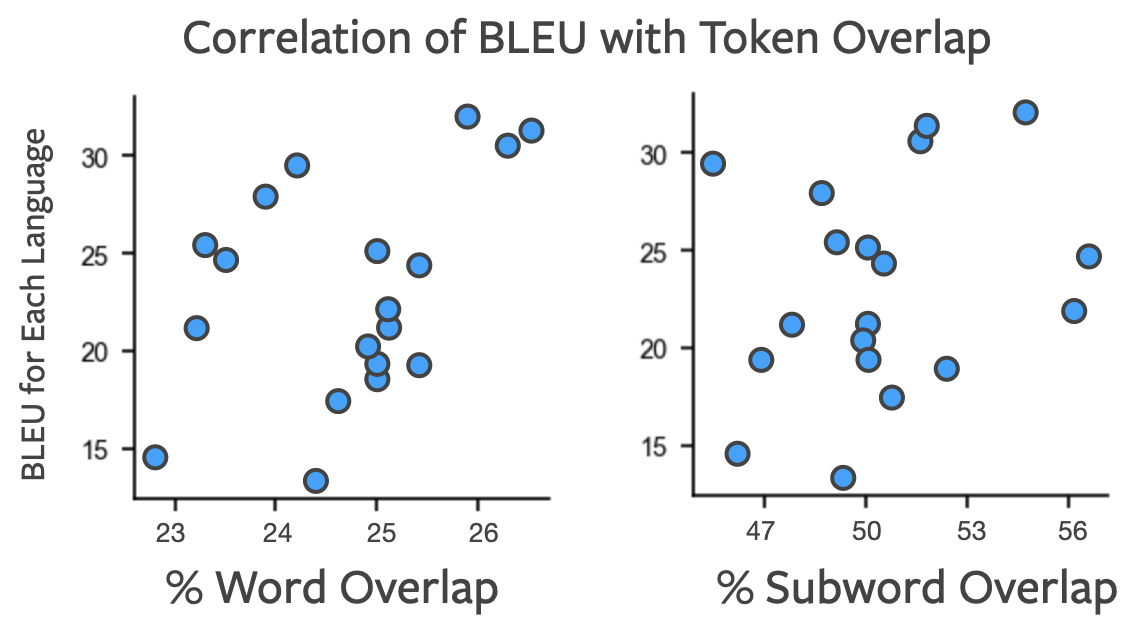}
    \caption{\textbf{Relationship between BLEU Score and Token Overlap} for all 21 languages. Correlation coefficient between word overlap and BLEU is 0.42, and coefficient between subword overlap and BLEU is 0.26.}
    \label{fig:type_token}
\end{figure}

Instead of operating on words, our models use sentencepiece \cite{wu2016google}, a data-driven approach to break 
words into subwords. As shown in \citet{wu-dredze-2019-beto}, in
transfer-based approaches to natural language understanding tasks, the proportion of subwords
shared between the source and the transfer language impacts
performance. We therefore explore the relation between the proportion of subwords and words shared between the AMR and the output vocabulary. Figure~\ref{fig:type_token} displays this relationship, with weak positive correlation for both word and subword overlap.

We further assess morphology by asking human evaluators to grade the
morphology of sentences (\textit{Is the morphology correct? Are agreement
constraints e.g., verb/subject, noun/adjective respected?}) on a
scale from 1 to 3 with 3 being the highest score. 
As Table~\ref{table:human_eval} shows, there is not much difference in performance between languages even
though there is a marked difference in terms of agreement constraints
between e.g., Finnish and English. Between annotators, agreement was high --- the standard deviation across  was low, with the exception of Romanian, Hungarian, and Spanish (as shown in Table~\ref{table:human_eval}). This demonstrates the
surprisingly high ability of multilingual models to
generalize across languages.

\paragraph{Word Order}
To assess the impact of varied word orders by language, we
ask human evaluators to judge if
the word order is natural. As shown in
Table~\ref{table:human_eval}, for all languages except Latvian and
Romanian, the score is very high (close to 3) indicating that the model
 learns to decode into multiple languages even though 
word order differs. The agreement between annotators was high, with low standard deviation (see Table~\ref{table:human_eval}). Further, the attention pattern between the
encoder English AMR and the decoder clearly
reflects the word order of the various languages. This is illustrated in Figure~\ref{fig:alignment}, where the activation pattern mirrors the word order difference between French (\ref{ex:frge-fr}) and German (\ref{ex:frge-ge}).

\enumsentence{
  \textit{ont tenu (une r\'{e}union de groupe)$\mathrm{_{OBJ}}$ (en Janvier 2020)$\mathrm{_{TIME}}$ (\`{a} New York)$\mathrm{_{LOC}}$}\\ \label{ex:frge-fr}\vspace{-0.75cm}
}
\enumsentence{{\textit{hielten (im Januar 2020)$\mathrm{_{TIME}}$ (in New York)$\mathrm{_{LOC}}$ (eine Gruppestreffen)$\mathrm{_{OBJ}}$} \label{ex:frge-ge}}
}

\paragraph{Training on Related Languages}

Multilingual models have the potential to benefit from similarities between languages. Languages of the same family often have shared morphological characteristics and vocabulary. First, we analyze the performance of training on languages within a family. Table~\ref{table:language_families} displays that a model trained on languages within a family has the strongest performance. 

\begin{figure}[t]
    \centering
    \includegraphics[width=\columnwidth]{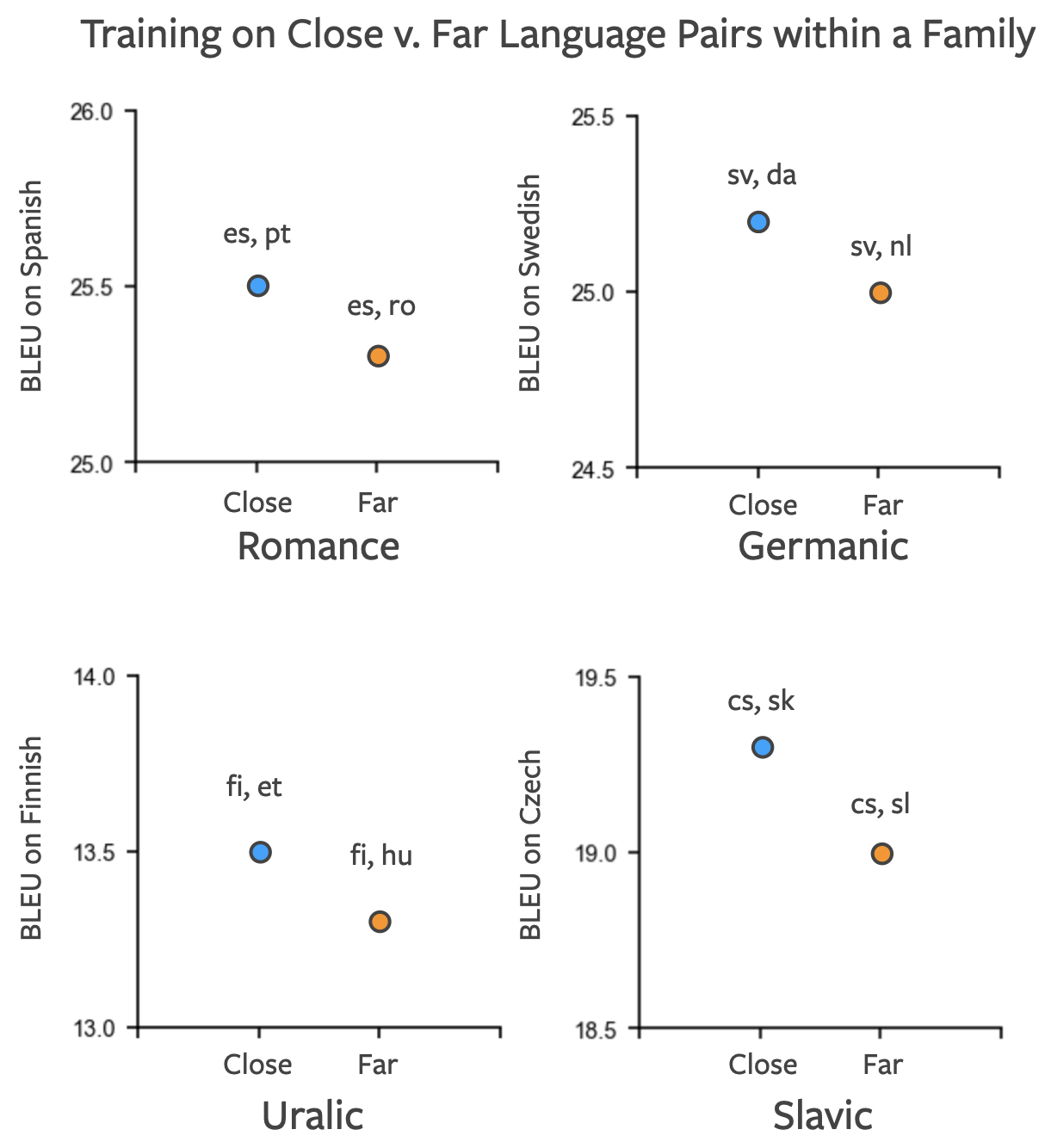}
    \caption{\textbf{BLEU difference training on Close v. Far Languages within One Family}. Training on a close pair consistently improves performance compared to training on a far pair, even within a language family.}
    \label{fig:pairs}
\end{figure}

Second, we analyze languages within the same family. For four families: Romance, Germanic, Uralic, and Slavic, we create multilingual models trained on pairs. One pair is for the most related languages within that family (e.g. Spanish and Portuguese) and another pair is for the farthest languages within that family (e.g. Spanish and Romanian). We determine which pairs are close and far from \citet{ahmad2019difficulties}. Results in Figure~\ref{fig:pairs} display that training on pairs of closely related languages has better performance than pairs of less closely related languages, even within a family. Multilingual models could pick up on similarities between languages to improve performance.

\paragraph{Semantic Accuracy and Paraphrasing.}
We ask human evaluators to grade the faithfulness of the hypothesis compared to the reference on a scale of 1 to 3. As shown in Table~\ref{table:human_eval}, the overall semantic accuracy is very high (note a score of 2 indicates \textit{minor differences}). We also asked annotators to evaluate how different the generated sentence was from the reference. When coupled with the semantic accuracy score, this allows us to evaluate generation of true paraphrases i.e., sentences with the same meaning as the reference but different surface form. In Table~\ref{table:human_eval}, \textit{Good Paraphrases} indicates the percentage of cases that scored highly (2 or 3) with respect to both semantic adequacy and paraphrasing.  
A large majority of generated sentences are labeled as valid paraphrases by native speakers, indicating (i) that despite underspecified input, the written sentence retains the meaning of the reference and (ii) that this underspecification allows for the generation of paraphrases. This also suggests that BLEU scores only partially reflect model performance as good paraphrases typically differ from the reference and are likely to get lower BLEU score even though they may be semantically accurate. Table~\ref{table:generation_examples} shows some examples illustrating the paraphrasing potential of the approach. 

\section{Conclusion}

Abstract Meaning Representations were designed to describe the meaning of English sentences. As such they are heavily biased towards English. AMR concepts are either English words, PropBank framesets (``want-01'') or special, English-based keywords (e.g., ``date-entity''). The structure of AMRs is also influenced by English syntax. For instance, the main relation of \textit{``I like to eat''} is the concept associated with its main verb (``like'') whereas given the corresponding German sentence \textit{"Ich esse gern'' (Lit. ``I eat willingly''}), the main predicate might have been chosen to be ``eat'' (``essen''). In other words, AMRs should not necessarily be viewed as an interlingua \cite{banarescu-etal-2013-abstract}. Nonetheless, our work suggests that it can be used as one: given an English-centric AMR it is possible to generate the corresponding sentence in multiple languages. This is in line with previous work by \cite{damonte_naacl18} which shows that despite translation divergences, AMR parsers can be learned for Italian, Chinese, German and Spanish which all map into an English-centric AMR. 

\section*{Acknowledgments}

We thank the anonymous reviewers for their feedback. We gratefully acknowledge the support of the French National Research Agency (Gardent; award ANR-20-CHIA-0003, XNLG "Multi-lingual, Multi-Source Text Generation"). 

\bibliographystyle{acl_natbib}
\bibliography{xling-amr2text,emnlp2020}

\clearpage 
\appendix 
\section{Appendix}

\subsection{Model Training Details}

For our baseline models, we train Transformer Big architectures with 240M parameters. We set the learning rate to $0.001$ with the inverse square root learning rate schedule from \citet{vaswani2017attention}, warming up for $4000$ updates. We train with the Adam optimizer with no weight decay and label smoothing $0.1$. We set the encoder and decoder layerdrop to $0.1$, and set the standard dropout to $0.3$ as well. We experiment with dropout values between $0.1, 0.2, 0.3$. We set the number of maximum tokens per batch to $3584$. We tune based on the validation loss at training time. We train for a fixed number of updates ($100,000$) and take the best checkpoint by validation loss. We train using 8 GPUs. The overall training time varies depending on the amount of training data available. Overall, we train for about a day and a half to two days, though good performance can be achieved within a day. The remaining training only marginally improves the quality as measured by BLEU.

For the multilingual models, we train with the same parameters as above, except the parameter size is slightly larger: around 250M parameters. The reason for this is the increased size of the XLM vocabulary that we use for initializing our cross-lingual embeddings. We again tune the dropout values between $0.1, 0.2, 0.3$. As we use pretraining and pretrained cross-lingual embeddings, we lower the learning rate to $0.0001$. A smaller learning rate can be used because the model parameters are initialized to a much better starting point. We warm up for $8000$ updates to ease the learning rate schedule at the beginning of training. We experimented with a variety of learning rates between $0.001-0.00001$, and tried five different values in this range. We chose the best performing value based on validation loss. The convergence speed is faster for multilingual models due to the pretraining initialization. Good performance can be achieved within half a day, though for experimental consistency we continue to train for the full $100,000$ updates to compare to the baseline. 

To generate from our models, we decode with beam search with beam size $5$. We experiment with beam size values between $4, 5$ and length penalty values between $0.4, 0.6, 1, 1.2$. We tune these values based on validation BLEU and use the best performing values to decode on the test set. Our decoding process is as follows: generate with the model on the validation set, remove the sentencepiece markers, then use the \texttt{sacrebleu} library for evaluation. As we use \texttt{sacrebleu}, we provide detokenized text.

\end{document}